\documentclass[runningheads]{llncs}

\usepackage{graphicx}
\usepackage{xcolor}
\usepackage{amssymb}
\usepackage{tabularx}
\usepackage{hyperref}

\usepackage{listings}
\usepackage{color, colortbl}
\definecolor{gray_1}{RGB}{142,152,178}
\definecolor{mygray}{gray}{0.95}

\newcommand{\toolname}{\textbf{LET}}

\begin{document}
\title{An Extensible Logic Embedding Tool for Lightweight Non-Classical Reasoning\\(system description)}

\titlerunning{An Extensible Logic Embedding Tool}

\author{
Alexander~Steen\orcidID{0000-0001-8781-9462}}

\authorrunning{A. Steen}

\institute{
  University of Greifswald, Germany
}

\maketitle              
\begin{abstract}
The logic embedding tool provides a procedural encoding for non-classical reasoning problems into
classical higher-order logic. It is extensible and can support an increasing number of different
non-classical logics as reasoning targets. When used as a pre-processor or library for
higher-order theorem provers, the tool admits off-the-shelf automation for logics for which otherwise
few to none provers are currently available.

\end{abstract}
\section{Introduction}
Non-classical logics (NCLs) deviate from various principles of classical logics
such as bivalence, truth-functionality, idempotency of entailment, etc.~\cite{Pri08}.
NCLs have numerous topical applications in artificial intelligence, mathematics, computer science, philosophy and other fields;
and increasingly many domain-specific NCLs are being introduced. 
Despite the relevance of NCL reasoning,
for many formalisms automated theorem proving (ATP) systems do not exist.
One major reason is that the development of ATP systems requires not only suitable theoretical foundations, but it also requires considerable
resources for software development and related aspects. It is not surprising that these efforts are only rarely 
made for logics that are still the subject of active research and discussion (i.e., moving targets),
and might be superseded with novel formalisms in the near future.
This situation impedes the deployment of methods in practical AI research, and it also hampers
the systematic evaluation of available formalisms.
Of course, there are notable exceptions of well-established NCLs for which ATP systems do exist,
such as linear logics~\cite{DBLP:conf/cade/ChaudhuriP05,DBLP:conf/tableaux/MantelO99},
intuitionistic logics~\cite{Ott21,Ott08,DBLP:conf/cade/SchmittLKN01,Tam97} and
modal logics~\cite{PN+21,Ott21,Ott14,TSK12,FF+01,HS00-TABLEAUX}.

Orthogonal to the development of special-purpose provers for individual NCLs is the use of logic translations that encode
the logic under consideration (the \emph{source logic}) into another logic formalism (the \emph{target logic})
for which there exist means of automation~\cite{Ohl91,Ohl93}. In this setting, improvements
to ATP systems for the target logic inherently benefit reasoning in the source logic.
A special type of logic translation is \emph{shallow embedding}~\cite{DBLP:conf/icfp/GibbonsW14}, in which the source logic's semantics is
directly encoded in the target logic.
 
The \emph{logic embedding tool} (\toolname) provides a library of shallow embeddings of NCLs into
higher-order logic,
and an executable for applying these embeddings on input problems. Special attention is paid to
the extensibility of the tool's underlying library of embeddings. \toolname\ is implemented in Scala
and freely available as open-source software (BSD-3 license) via Zenodo~\cite{Ste22-LE} and GitHub\footnote{
\href{https://github.com/leoprover/logic-embedding}{\texttt{github.com/leoprover/logic-embedding}}
}.
The input format is a non-classical TPTP syntax extension~\cite{SF+22} that allows 
non-classical reasoning problems to be written within the common TPTP framework~\cite{Sut17}, see Sect.~\ref{sec:syntax}
for an overview.
\toolname\ can be used as library or as external pre-processor to higher-order ATP systems, effectively enabling
automated reasoning for various NCLs. Currently, the following logics are supported:
\begin{itemize}
  \item Many quantified normal multi-modal logics
  \item Various hybrid logics
  \item Public announcement logic
  \item Carmo and Jones' dyadic deontic logic
  \item Åqvist's dyadic deontic logic \textbf{E}
\end{itemize}

\section{Problem Representation Format \label{sec:syntax}}

As input syntax \toolname\ accepts the TFN and THN languages~\cite{SF+22}, recent non-classical extensions of the well-established
TPTP syntax standard for ATP systems. 
The TPTP syntax is part of the TPTP World infrastructure~\cite{Sut17}, and defines several languages for representing
reasoning problems and solutions, including languages for untyped first-order logic (FOF)~\cite{Sut09},
typed first order logic (TFF)~\cite{SS+12,BP13-TFF1},
typed first-order logic extended with Boolean terms and variables (TXF, formerly TFX)~\cite{SK18},
and higher-order logic (THF)~\cite{SB10,KSR16}. A comprehensive survey of these languages and their usage
is available in the literature~\cite{Sut17}.

The TXN and THN languages extend TXF and THF, respectively, with
new generic non-classical operators of the form
{\tt \verb|{|}{\em connective\_name}{\tt \verb|}|} that are
applied like function symbols. For some operators there are also short forms available (not discussed here).
Although TXN is a typed language, it may also represent untyped formalisms. Following the conventions
from TFF and TXF, any predicate symbol and function symbol in the problem with undeclared type implicitly defaults to
a canonical $n$-ary predicate type or $n$-ary function type.

Additionally, TXN and THN introduce so-called \emph{logic specifications},
special kinds of TPTP annotated formulas with the role {\tt logic}, which specify the logic being used within the problem file.
In TXN they are of form \ldots\\
\hspace*{1cm}{\tt tff(}{\em name}{\tt,logic,}{\em logic\_name} {\tt ==} {\em properties}{\tt ).} \\
where {\em logic\_name} is some TPTP or user defined name for a logic (or logic family), and
{\em properties} is a list of key-value parameters that optionally further specify the intended
NCL. In THN the format is the same, only that the THF formula identifier {\tt thf} is used instead.

A detailed introduction of non-classical connectives and logic specifications is presented in the respective TPTP
proposal~\cite{tptpNCL,SF+22}, and they are informally illustrated via the application examples below.

\section{Architecture}

\begin{figure}[t]
  \centering
  \includegraphics[width=0.98\textwidth,interpolate]{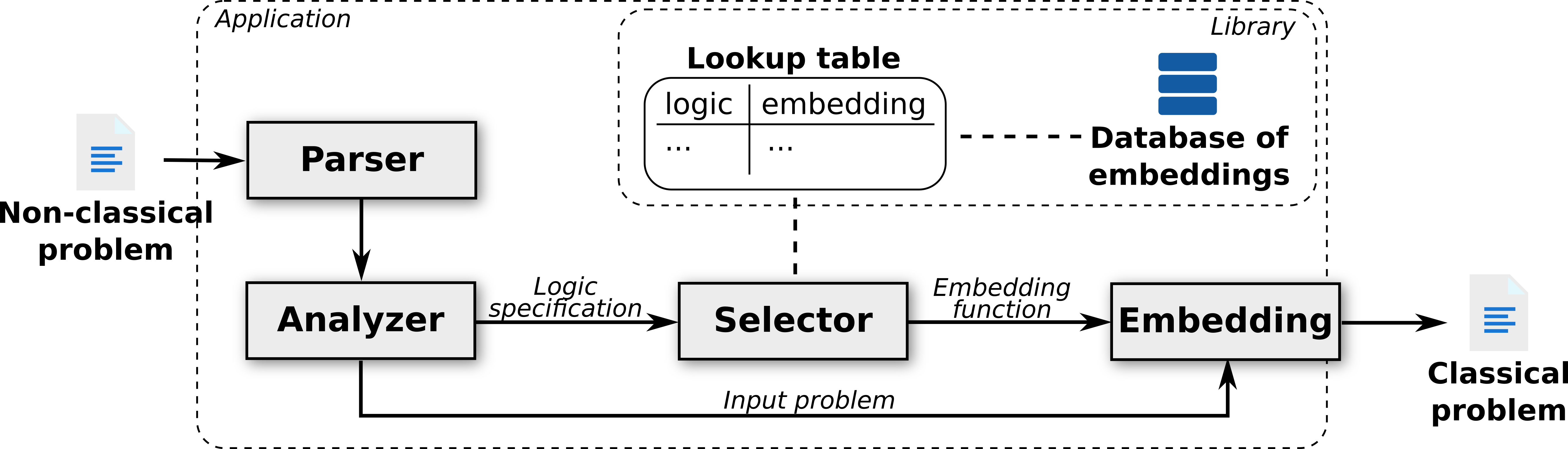}
  \caption{The top-level architecture of \toolname. The solid arrows indicate directed data flow between
  different components, the dashed lines represent access to additional resources. \label{fig:process}}
\end{figure}

The components of \toolname\ and their relationship are displayed in Fig.~\ref{fig:process}.
It is structured into two main modules:
\begin{enumerate}
  \item The \emph{library} module defines a common embedding interface, and constitutes the collection of shallow embeddings for
        different NCLs. 
  \item The \emph{application} module implements a stand-alone executable on top of the library module.
        It finds and applies the correct shallow embedding on a given input problem.
\end{enumerate}
Note that the library module is independent from the application module, and can be included in existing ATP systems
via a simple API. The application module, in contrast, can be employed as external pre-processor executable.

The general procedure implemented by the application module is as follows:
\begin{enumerate}
\item The input problem is parsed~\cite{Ste21} and scanned for a logic specification
      (an annotated TPTP formula with the role \texttt{logic}),
\item the logic name and the parameters are extracted from the logic specification,
\item the database of supported logics is queried for the given logic and, if supported,
      the respective embedding procedure is provided, and finally,
\item the embedding procedure is invoked on the input problem and the result is returned
      as classical TPTP THF problem.
\end{enumerate}
If the problem does not contain any logic specification, the original problem is returned.
If the logic specified in the input problem is malformed or not supported by \toolname\ an error is reported.
The separation of library and application facilitates \toolname's extensibility, as the library can be easily
extended with new embeddings of further NCLs while the application module remains unchanged.

Note that the output of the tool is a classical higher-order problem represented in THF syntax.
Hence every higher-order ATP system that supports reasoning in THF can be employed for reasoning in the respective NCL.
Additionally, \toolname\ supports TSTP-compatible result reporting~\cite{SZS04} for seamless integration into TPTP/TSTP tool chains.

\section{Overview of Supported NCLs}
The NCLs currently supported by \toolname\ are the following:
\begin{description}
  \item[Modal logics.] The logic name \verb|$modal| represents the family of
  propositional and first-order quantified normal multi-modal
  logics~\cite{DBLP:books/el/07/BlackburnB07,BG07}. The modal operators
  $\Box$ and $\Diamond$ are represented by the non-classical connectives \verb|{$box}|
  and \verb|{$dia}|, respectively. In the case of multiple modalities, the
  connectives are indexed with uninterpreted user constants, prefixed with a 
  \verb|#| (hash sign) as, e.g., in \verb|{$box(#i)}| and \verb|{$dia(#i)}|.
  Global assumptions via the role \texttt{axiom}, and
  local assumptions are expressed via annotated formulas of role \texttt{hypothesis}~\cite{FM98}. 
  Relevant embeddings are described in~\cite{BP13,GSB17}.
  \item[Hybrid logics.] Hybrid logics, referred to as \verb|$$hybrid|,
    extend the modal logic family \verb|$modal| with
    the notion of nominals, a special kind of atomic formula symbol
    that is true only in a specific world~\cite{BBW07}.
    The logics represented by \verb|$$hybrid| are first-order variants
    of $\mathcal{H}(\mathsf{E},@, \downarrow)$~\cite{BBW07,BG07}.
    A nominal symbol $n$ is represented as \verb|{$$nominal}(n)|,
    the shift operator $@_s$ as \verb|{$$shift(#s)}|,
    and the bind operator $\downarrow x$ as \verb|{$$bind(#X)}|.
    All other aspects are analogous to the modal logic representation above.
  Preliminary shallow embeddings results are reported in~\cite{WS14}.
  The embedding implemented in \toolname\ simplifies and extends these.
  \item[Public announcement logic.] Public announcement logic (PAL), \verb|$$pal|,
    is a propositional epistemic logic that allows for reasoning about knowledge.
    In contrast to \verb|$modal|, PAL is a dynamic logic 
    that supports updating the knowledge of agents via so-called announcement
    operators~\cite{DBLP:journals/corr/DitmarschHHK15}.
    The knowledge operator $K_i$ is given by \verb|{$$knows(#i)}|,
    the common knowledge operator $C_A$, with $A$ a set of agents,
    by \verb|{$$common($$group := [...])}|, and the
    announcement $[!\varphi]$ is represented as
    \verb|{$$announce($$formula := phi)}|.
    An embedding of PAL is presented in~\cite{DBLP:journals/afp/BenzmullerR21}.
  \item[Dyadic deontic logics.] Deontic logics are formalisms for reasoning over norms, obligations, permissions and prohibitions.
    In contrast to modal logics used for this purpose (e.g., modal logic \textbf{D}),
    dyadic deontic logics (DDLs), named \verb|$$ddl|, offer a more sophisticated representation of conditional norms using
    a dyadic obligation operator $\bigcirc(\varphi/\psi)$. They address paradoxes of other deontic
    logics in the context of so-called contrary-to-duty (CTD) situations~\cite{chisholm1963contrary}.
    The concrete DDLs supported by \toolname\ are the propositional system by Carmo and Jones~\cite{CJ13}
    and Åqvist's propositional system \textbf{E}~\cite{aaqvist2002deontic}.
    The dyadic deontic operator $\bigcirc$ is represented by
    \verb|{$$obl}| (short for obligatory).
    An embedding of the above DDL is studied in~\cite{BFP22,BFP19}
\end{description}
Note that the name \verb|$modal| of modal logics and that of its connective names are given by
TPTP defined names (starting with a single dollar sign) since it is
the first non-classical logic standardized by the TPTP~\cite{tptpNCL,SF+22}.
All further logics are \toolname-specific logic representations that have not (yet) been
included in the collection of TPTP curated NCLs; following the TPTP naming 
convention, their identifiers hence start with two dollar signs (system defined names).
  
Non-classical logic languages quite commonly admit different concrete logics using the same
syntax. In order to choose the exact logic intended for the input problem, suitable parameters are
given as properties to the logic specification as introduced in Sect.~\ref{sec:syntax}.
For the above NCLs supported by \toolname, Table~\ref{table:logics} gives an overview of the
individual parameters and their meaning. We refer to Fitting and Mendelsohn~\cite{FM98} for an explanation
of the modal logic properties.

\begin{table}[t]
\caption{Overview of the parameters of the different NCLs supported by \toolname \label{table:logics}}
     \centering
     \begin{tabularx}{0.98\textwidth}{ l | c | X }
         \multicolumn{1}{c|}{\bfseries Logic} &
         \multicolumn{1}{c|}{\bfseries Parameter} &
         \multicolumn{1}{c}{\bfseries Description}\\
         \hline
         \verb|$modal|   &
           \verb|$quantification| & \parbox[t]{.64\textwidth}{%
                                    Selects whether quantification semantics is varying domains,
                                    constant domains, cumulative domains or decreasing domains.
                                    
                                    \smallskip
                                    Accepted values: \verb|$varying|, \verb|$constant|, \verb|$cumulative|,
                                    \verb|$decreasing|
                                    } \\[1.5em]
          & \cellcolor{mygray}{
           \verb|$constants|} & \cellcolor{mygray} 
           \parbox[t]{.64\textwidth}{
                                    Selects whether constant and functions symbols are interpreted as rigid
                                    or flexible.
                                    
                                    \smallskip
                                    Accepted values: \verb|$rigid|, \verb|$flexible|$^\dagger$\\[.5em]
                                    {\scriptsize $\dagger$: Not yet supported by \toolname.}
                                    } \\[1.5em]
          &
           \verb|$modalities| & 
           \parbox[t]{.64\textwidth}{
                                    Selects the properties for the modal operators.
                                    
                                    \smallskip
                                    Accepted values, for each modality:\\
                                    \verb|$modal_system_X| where \texttt{X} $\in$ 
                                    \{{\tt K}, {\tt KB}, {\tt K4}, {\tt K5}, {\tt K45}, {\tt KB5}, {\tt D},
                                      {\tt DB}, {\tt D4}, {\tt D5}, {\tt D45}, {\tt T},
                                      {\tt B}, {\tt S4}, {\tt S5}, {\tt S5U}\} \\
                                      {\em or a list of axiom schemes} \\
                        {\tt [\$modal\_axiom\_X$_1$, \ldots, \$modal\_axiom\_X$_n$]} \\
                        {\tt X$_i$} $\in$ \{{\tt K}, {\tt T}, {\tt B}, {\tt D}, {\tt 4}, {\tt 5}, 
                         {\tt CD}, {\tt C4}\}
                                    } \\
         \rowcolor{mygray}
         \verb|$$hybrid| & \textit{see} \verb|$modal| & \textit{see} \verb|$modal| \\[.2em]
         \verb|$$pal|    & \emph{none}    & \emph{---} \\[.2em]
         \rowcolor{mygray}
         \verb|$$ddl|    &
          \verb|$$system| & \parbox[t]{.64\textwidth}{
                                    Selects which DDL logic system is employed: Carmo and Jones or Åqvist's system \textbf{E}.
                                    
                                    \smallskip
                                    Accepted values: \verb|$$carmoJones| or \verb|$$aqvistE|
                                    }
     \end{tabularx}
\end{table}

\section{Application Examples}
The functionality of \toolname\ is illustrated by a number of examples. 
Exemplary ATP system results are produced by the higher-order prover Leo-III~\cite{SB21}, version 1.6.8,
in which \toolname\ is integrated as a library and accessed via its API. Leo-III parses the problems,
invokes the embedding API, and then applies standard proof search on the resultant THF problem.

\paragraph{Example 1: Modal logic reasoning.}
The Barcan formula~\cite{Bar46}, given by
$$\forall x.\, \Box p(x) \Rightarrow \Box (\forall x.\, p(x)) $$
in a first-order
variant, is a modal logic formula that is valid if and only if the quantification domain of the underlying first-order
modal logic model is non-cumulative~\cite{FM98}. This is written in TXN as \ldots

\begin{lstlisting}[frame=single,basicstyle=\ttfamily\small,columns=fullflexible,keepspaces=true]
  tff(modal_k5, logic, $modal == [
     $constants == $rigid,
     $quantification == $decreasing,
     $modalities == [$modal_axiom_K, $modal_axiom_5]
   ] ).

  tff(bf, conjecture, ( ![X]: ({$box}(f(X))) ) => {$box}(![X]: f(X)) ).
\end{lstlisting}
This specifies a modal logic with rigid function symbols, decreasing quantification domains and 
box operators satisfying modal axiom schemes $K$ and $5$. As expected, Leo-III returns \ldots
\begin{lstlisting}[basicstyle=\ttfamily\small,columns=fullflexible,keepspaces=true]
  % SZS status Theorem for barcan.p
\end{lstlisting}
However, when the parameter \verb|$quantification| is changed to \verb|$cumulative| or \verb|$varying|
the problem becomes countersatisfiable.

\paragraph{Example 2: Hybrid logic reasoning.}
Hybrid logics can talk about the satisfaction relation of the modal logic at the object language level.
Up to the author's knowledge, Leo-III is the first ATP system to support reasoning in many different first-order hybrid logics.
An example tautology is given by 
$$ \forall X.\, \Box @_n \big(\downarrow \! Y.\, (Y \land p(X)) \Leftrightarrow (n \land p(X))\big)$$
that is encoded as \ldots
\begin{lstlisting}[frame=single,basicstyle=\ttfamily\small,columns=fullflexible,keepspaces=true]
  tff(hybrid_s5,logic, $$hybrid == [
      $constants == $rigid,
      $quantification == $varying,
      $modalities == $modal_system_S5
    ] ).

  tff(1, conjecture, ![X]: {$box}({$$shift(#n)}(
             {$$bind(#Y)}((Y & p(X))
                          <=> ({$$nominal}(n) & p(X)) ))) ).
\end{lstlisting}

\paragraph{Example 3: CTD reasoning in deontic logics.}
In deontic logics, CTD situations arise when reasoning with obligations that prescribe what to do if other (primary) obligations
are violated. Simple approaches, e.g., using modal logic \textbf{D}, lead to inconsistencies that
allow arbitrary conclusions to be inferred. This is addressed by dyadic deontic logics that encode conditional norms
using a special operator $\bigcirc(\psi,\varphi)$ (read: \emph{it ought to be $\psi$, given $\varphi$}). An example is \ldots

\begin{lstlisting}[frame=single,basicstyle=\ttfamily\small,columns=fullflexible,keepspaces=true]
  tff(spec_e, logic, $$ddl == [ $$system == $$aqvistE ] ).

  tff(a1, axiom, {$$obl}(go,$true)).
  tff(a2, axiom, {$$obl}(tell, go)).
  tff(a3, axiom, {$$obl}(~tell, ~go)).
  tff(situation, axiom, ~go). 
  tff(c, conjecture, {$$obl}(~tell,$true)).
\end{lstlisting}
This example encodes that (a1) you ought to go and help your neighbors, (a2) if you go then you ought to tell them that you are coming,
and (a3) if you don't go, then you ought not tell them. It can consistently be inferred that if you actually don't go, then you ought not
tell them; Leo-III confirms this. Up to the author's knowledge, Leo-III is the first ATP system to support system \textbf{E}
and the system of Carmo and Jones.

\section{Summary}
\toolname\ is a library and pre-processing tool for encoding non-classical reasoning problems
into classical higher-order logic, by means of shallow embeddings. The output of the tool is TPTP THF, and any compatible ATP system
can be used in conjunction with it, offering of-the-shelf automation for non-classical logics.
Although the range of supported logic families is still quite limited, already at this point \toolname\
allows for the automation of more than 60 different first-order modal logics (including all logics from the modal cube),
60 different hybrid logics, dynamic epistemic logic (PAL), and different dyadic deontic logics.\footnote{
The number of modal logics is at least $15~\textit{(modality axiomatizations)} \times 4~\textit{(quantification semantics)} = 60$.
Many more modal logics are supported since \toolname\ allows arbitrary
combinations of different modalities. 
Also, quantification semantics can be controlled on a per-type basis.
}
For some of these logics there exist no other ATP systems to date. The tool is designed to be
easily extensible with new embeddings of further logics.
Shallow semantical embeddings into HOL have also been studied for various other purposes~\cite{Ben19}.

Shallow embeddings and hence \toolname\ target rapid logic prototyping,
but might not be as
effective as ATP systems specifically designed for the respective NCL.
The embedding approach allows for the automation of logics
that otherwise would have no automation at all. \toolname\ aims at closing automation gaps
for interesting NCLs, rather than challenging available ATP systems.
Nevertheless, previous studies indicate that embeddings perform quite competitively in the context of quantified benchmarks~\cite{T3}.
However, for many of the logics currently covered by \toolname, in particular for quantified logics, there are no benchmark sets or competitors
available, and comparisons are not possible. Automation via embeddings can also be employed in an educational context for
low-threshold student experiments.

\toolname\ generalizes and extends previous work on modal logic embedding tools \cite{GSB17,GS18},
and offers more encoding variants for modal logics, including embedding into polymorphic THF (not discussed in detail here).
\toolname\ makes use of the novel non-classical TPTP format and is seamlessly included into the Leo-III prover
so that no extra steps are necessary for non-classical reasoning.

\bibliographystyle{splncs04}
\bibliography{Bibliography}

\begin{thebibliography}{10}
\providecommand{\url}[1]{\texttt{#1}}
\providecommand{\urlprefix}{URL }
\providecommand{\doi}[1]{https://doi.org/#1}

\bibitem{aaqvist2002deontic}
{\AA}qvist, L.: Deontic logic. In: Handbook of philosophical logic, pp.
  147--264. Springer (2002)

\bibitem{BBW07}
Areces, C., ten Cate, B.: Hybrid logics. In: Blackburn, P., van Benthem,
  J.F.A.K., Wolter, F. (eds.) Handbook of Modal Logic, Studies in logic and
  practical reasoning, vol.~3, pp. 821--868. North-Holland (2007)

\bibitem{Bar46}
Barcan, R.: {A Functional Calculus of First Order Based on Strict Implication}.
  Journal of Symbolic Logic  \textbf{11},  1--16 (1946)

\bibitem{Ben19}
Benzm{\"u}ller, C.: {Universal (Meta-)Logical Reasoning: Recent Successes}.
  Science of Computer Programming  \textbf{172},  48--62 (2019)

\bibitem{BFP19}
Benzm{\"u}ller, C., Farjami, A., Parent, X.: {{\AA}qvist's Dyadic Deontic Logic
  {E} in {HOL}}. Journal of Applied Logics  \textbf{6}(5),  733--755 (2019)

\bibitem{BP13}
Benzm{\"u}ller, C., Paulson, L.: {Quantified Multimodal Logics in Simple Type
  Theory}. Logica Universalis  \textbf{7}(1),  7--20 (2013)

\bibitem{BFP22}
Benzm{\"u}ller, C., Farjami, A., Parent, X.: Dyadic deontic logic in {HOL}:
  Faithful embedding and meta-theoretical experiments. In: New Developments in
  Legal Reasoning and Logic, pp. 353--377. Springer (2022)

\bibitem{DBLP:journals/afp/BenzmullerR21}
Benzm{\"{u}}ller, C., Reiche, S.: {Automating Public Announcement Logic and the
  Wise Men Puzzle in Isabelle/HOL}. Arch. Formal Proofs  \textbf{2021} (2021),
  \url{https://www.isa-afp.org/entries/PAL.html}

\bibitem{DBLP:books/el/07/BlackburnB07}
Blackburn, P., van Benthem, J.: Modal logic: a semantic perspective. In:
  Blackburn, P., van Benthem, J.F.A.K., Wolter, F. (eds.) Handbook of Modal
  Logic, Studies in logic and practical reasoning, vol.~3, pp. 1--84.
  North-Holland (2007)

\bibitem{BP13-TFF1}
Blanchette, J., Paskevich, A.: {TFF1: The TPTP Typed First-order Form with
  Rank-1 Polymorphism}. In: Bonacina, M. (ed.) {Proceedings of the 24th
  International Conference on Automated Deduction}. pp. 414--420. No.~7898 in
  Lecture Notes in Artificial Intelligence, Springer-Verlag (2013)

\bibitem{BG07}
Bra{\"{u}}ner, T., Ghilardi, S.: First-order modal logic. In: Blackburn, P.,
  van Benthem, J.F.A.K., Wolter, F. (eds.) Handbook of Modal Logic, Studies in
  logic and practical reasoning, vol.~3, pp. 549--620. North-Holland (2007)

\bibitem{CJ13}
Carmo, J., Jones, A.J.I.: Completeness and decidability results for a logic of
  contrary-to-duty conditionals. J. Log. Comput.  \textbf{23}(3),  585--626
  (2013)

\bibitem{FF+01}
Fari{\~n}as~del Cerro, L., Fauthoux, D., Gasquet, O., Herzig, A., Longin, D.,
  Massacci, F.: {LoTREC: The Generic Tableau Prover for Modal and Description
  Logics}. In: Gore, R., Leitsch, A., Nipkow, T. (eds.) {Proceedings of the
  International Joint Conference on Automated Reasoning}. pp. 453--458.
  No.~2083 in Lecture Notes in Artificial Intelligence, Springer-Verlag (2001)

\bibitem{DBLP:conf/cade/ChaudhuriP05}
Chaudhuri, K., Pfenning, F.: A focusing inverse method theorem prover for
  first-order linear logic. In: Nieuwenhuis, R. (ed.) Proceedings of the 20th
  International Conference on Automated Deduction. Lecture Notes in Computer
  Science, vol.~3632, pp. 69--83. Springer (2005)

\bibitem{chisholm1963contrary}
Chisholm, R.M.: Contrary-to-duty imperatives and deontic logic. Analysis
  \textbf{24}(2),  33--36 (1963)

\bibitem{DBLP:journals/corr/DitmarschHHK15}
van Ditmarsch, H., Halpern, J.Y., van~der Hoek, W., Kooi, B.P.: An introduction
  to logics of knowledge and belief. In: Handbook of epistemic logic (2015)

\bibitem{FM98}
Fitting, M., Mendelsohn, R.: {First-Order Modal Logic}. Kluwer (1998)

\bibitem{DBLP:conf/icfp/GibbonsW14}
Gibbons, J., Wu, N.: Folding domain-specific languages: deep and shallow
  embeddings (functional pearl). In: Jeuring, J., Chakravarty, M.M.T. (eds.)
  Proceedings of the 19th {ACM} {SIGPLAN} international conference on
  Functional programming. pp. 339--347. {ACM} (2014)

\bibitem{GS18}
Glei{\ss}ner, T., Steen, A.: {The MET: The Art of Flexible Reasoning with
  Modalities}. In: Benzm{\"u}ller, C., Ricca, F., Parent, X., Roman, D. (eds.)
  {Proceedings of the 2nd International Joint Conference on Rules and
  Reasoning}. pp. 274--284. No. 11092 in Lecture Notes in Computer Science
  (2018)

\bibitem{GSB17}
Glei{\ss}ner, T., Steen, A., Benzm{\"u}ller, C.: {Theorem Provers for Every
  Normal Modal Logic}. In: Eiter, T., Sands, D. (eds.) {Proceedings of the 21st
  International Conference on Logic for Programming, Artificial Intelligence,
  and Reasoning}. pp. 14--30. No.~46 in EPiC Series in Computing, EasyChair
  Publications (2017)

\bibitem{tptpNCL}
Glei{\ss}ner, T., Steen, A., Sutcliffe, G., Benzm{\"{u}}ller, C.: {TPTP}
  proposal: Non-classical logics. http://tptp.org/NonClassicalLogic (2021)

\bibitem{HS00-TABLEAUX}
Hustadt, U., Schmidt, R.: {MSPASS: Modal Reasoning by Translation and
  First-Order Resolution}. In: Dyckhoff, R. (ed.) {Proceedings of the
  International Conference on Automated Reasoning with Analytic Tableaux and
  Related Methods}. pp. 67--71. No.~1847 in Lecture Notes in Artificial
  Intelligence, Springer-Verlag (2000)

\bibitem{KSR16}
Kaliszyk, C., Sutcliffe, G., Rabe, F.: {TH1: The TPTP Typed Higher-Order Form
  with Rank-1 Polymorphism}. In: Fontaine, P., Schulz, S., Urban, J. (eds.)
  {Proceedings of the 5th Workshop on Practical Aspects of Automated
  Reasoning}. pp. 41--55. No.~1635 in CEUR Workshop Proceedings (2016)

\bibitem{DBLP:conf/tableaux/MantelO99}
Mantel, H., Otten, J.: lintap: {A} tableau prover for linear logic. In: Murray,
  N.V. (ed.) Proceedings of the 8th Conference on Automated Reasoning with
  Analytic Tableaux and Related Methods. Lecture Notes in Computer Science,
  vol.~1617, pp. 217--231. Springer (1999)

\bibitem{Ohl91}
Ohlbach, H.: {Semantics Based Translation Methods for Modal Logics}. Journal of
  Logic and Computation  \textbf{1}(5),  691--746 (1991)

\bibitem{Ohl93}
Ohlbach, H.: {Translation Methods for Non-Classical Logics: An Overview}. Logic
  Journal of the IGPL  \textbf{1}(1),  69--89 (1993)

\bibitem{Ott08}
Otten, J.: {leanCoP 2.0 and ileancop 1.2: High Performance Lean Theorem Proving
  in Classical and Intuitionistic Logic}. In: Baumgartner, P., Armando, A.,
  Dowek, G. (eds.) {Proceedings of the 4th International Joint Conference on
  Automated Reasoning}. pp. 283--291. No.~5195 in Lecture Notes in Artificial
  Intelligence (2008)

\bibitem{Ott14}
Otten, J.: {MleanCoP: A Connection Prover for First-Order Modal Logic}. In:
  Demri, S., Kapur, D., Weidenbach, C. (eds.) {Proceedings of the 7th
  International Joint Conference on Automated Reasoning}. pp. 269--276.
  No.~8562 in Lecture Notes in Artificial Intelligence (2014)

\bibitem{Ott21}
Otten, J.: {The nanoCoP 2.0 Connection Provers for Classical, Intuitionistic
  and Modal Logics}. In: Das, A., Negri, S. (eds.) {Proceedings of the 30th
  International Conference on Automated Reasoning with Analytic Tableaux and
  Related Methods}. pp. 236--249. No. 12842 in Lecture Notes in Artificial
  Intelligence, Springer-Verlag (2021)

\bibitem{PN+21}
Papacchini, F., Nalon, C., Hustadt, U., Dixon, C.: {Efficient Local Reductions
  to Basic Modal Logic}. In: Platzer, A., Sutcliffe, G. (eds.) {Proceedings of
  the 28th International Conference on Automated Deduction}. pp. 76--92. No.
  12699 in Lecture Notes in Computer Science, Springer-Verlag (2021)

\bibitem{Pri08}
Priest, G.: {An Introduction to Non-Classical Logic: From If to Is}. Cambridge
  University Press (2008)

\bibitem{DBLP:conf/cade/SchmittLKN01}
Schmitt, S., Lorigo, L., Kreitz, C., Nogin, A.: Jprover: Integrating
  connection-based theorem proving into interactive proof assistants. In:
  Gor{\'{e}}, R., Leitsch, A., Nipkow, T. (eds.) Proceedings of the First
  International Joint Conference on Automated Reasoning. Lecture Notes in
  Computer Science, vol.~2083, pp. 421--426. Springer (2001)

\bibitem{Ste22-LE}
Steen, A.: {logic-embedding v1.7} (2022), {DOI: 10.5281/zenodo.5913215}

\bibitem{Ste21}
Steen, A.: {Scala TPTP Parser v1.6} (2022), {DOI: 10.5281/zenodo.4468958}

\bibitem{SB21}
Steen, A., Benzm{\"u}ller, C.: {Extensional Higher-Order Paramodulation in
  Leo-III}. Journal of Automated Reasoning  \textbf{65}(6),  775–807 (2021)

\bibitem{SF+22}
Steen, A., Fuenmayor, D., Glei{\ss}ner, T., Sutcliffe, G., Benzm{\"u}ller, C.:
  {Automated Reasoning in Non-classical Logics in the TPTP World}. In:
  Blanchette, J., Kovacs, L., Pattinson, D. (eds.) {Proceedings of the 11th
  International Joint Conference on Automated Reasoning}. p. Confidently
  submitted. Lecture Notes in Artificial Intelligence (2022), {Preprint
  available at \url{https://arxiv.org/abs/2202.09836}}

\bibitem{T3}
Steen, A.: Extensional Paramodulation for Higher-Order Logic and its Effective
  Implementation Leo-III, {DISKI} -- Dissertations in Artificial Intelligence,
  vol.~345. Akademische Verlagsgesellschaft AKA GmbH, Berlin (9 2018),
  {Dissertation}, Freie Universit{\"a}t Berlin, Germany.

\bibitem{Sut09}
Sutcliffe, G.: {The TPTP Problem Library and Associated Infrastructure. The FOF
  and CNF Parts, v3.5.0}. Journal of Automated Reasoning  \textbf{43}(4),
  337--362 (2009)

\bibitem{Sut17}
Sutcliffe, G.: {The TPTP Problem Library and Associated Infrastructure. From
  CNF to TH0, TPTP v6.4.0}. Journal of Automated Reasoning  \textbf{59}(4),
  483--502 (2017)

\bibitem{SB10}
Sutcliffe, G., Benzm{\"u}ller, C.: {Automated Reasoning in Higher-Order Logic
  using the TPTP THF Infrastructure}. Journal of Formalized Reasoning
  \textbf{3}(1),  1--27 (2010)

\bibitem{SK18}
Sutcliffe, G., Kotelnikov, E.: {TFX: The TPTP Extended Typed First-order Form}.
  In: Konev, B., Urban, J., Schulz, S. (eds.) {Proceedings of the 6th Workshop
  on Practical Aspects of Automated Reasoning}. pp. 72--87. No.~2162 in CEUR
  Workshop Proceedings (2018)

\bibitem{SS+12}
Sutcliffe, G., Schulz, S., Claessen, K., Baumgartner, P.: {The TPTP Typed
  First-order Form with Arithmetic}. In: Bj{\o}rner, N., Voronkov, A. (eds.)
  {Proceedings of the 18th International Conference on Logic for Programming,
  Artificial Intelligence, and Reasoning}. pp. 406--419. No.~7180 in Lecture
  Notes in Artificial Intelligence, Springer-Verlag (2012)

\bibitem{SZS04}
Sutcliffe, G., Zimmer, J., Schulz, S.: {TSTP Data-Exchange Formats for
  Automated Theorem Proving Tools}. In: Zhang, W., Sorge, V. (eds.)
  {Distributed Constraint Problem Solving and Reasoning in Multi-Agent
  Systems}, pp. 201--215. No.~112 in Frontiers in Artificial Intelligence and
  Applications, IOS Press (2004)

\bibitem{Tam97}
Tammet, T.: {Gandalf}. Journal of Automated Reasoning  \textbf{18}(2),
  199--204 (1997)

\bibitem{TSK12}
Tishkovsky, D., Schmidt, R., Khodadadi, M.: {The Tableau Prover Generator
  MetTeL2}. In: Platzer, A., Sutcliffe, G. (eds.) {Proceedings of the 13th
  European conference on Logics in Artificial Intelligence}. pp. 492--495.
  No.~7519 in Lecture Notes in Computer Science, Springer (2012)

\bibitem{WS14}
Wisniewski, M., Steen, A.: Embedding of quantified higher-order nominal modal
  logic into classical higher-order logic. In: Benzm{\"{u}}ller, C., Otten, J.
  (eds.) Workshop on Automated Reasoning in Quantified Non-Classical Logics.
  EPiC Series in Computing, vol.~33, pp. 59--64. EasyChair (2014)

\end{thebibliography}

\end{document}